\documentclass[acmtog,authorversion,nonacm]{acmart}
\usepackage{booktabs} 

\usepackage{xcolor}  
\usepackage{hyperref} 

\usepackage{multirow}
\usepackage[ruled]{algorithm2e} 

\SetAlFnt{\small}
\SetAlCapFnt{\small}
\SetAlCapNameFnt{\small}
\SetAlCapHSkip{0pt}
\usepackage{float} 
\acmJournal{TOG}
\usepackage{graphicx}
\usepackage{booktabs} 
\usepackage{array}

\def\mytitle{VFX Creator: Animated Visual Effect Generation with Controllable Diffusion Transformer}

\begin{document}

\begin{teaserfigure}
\centering
\includegraphics[width=1.0\textwidth]{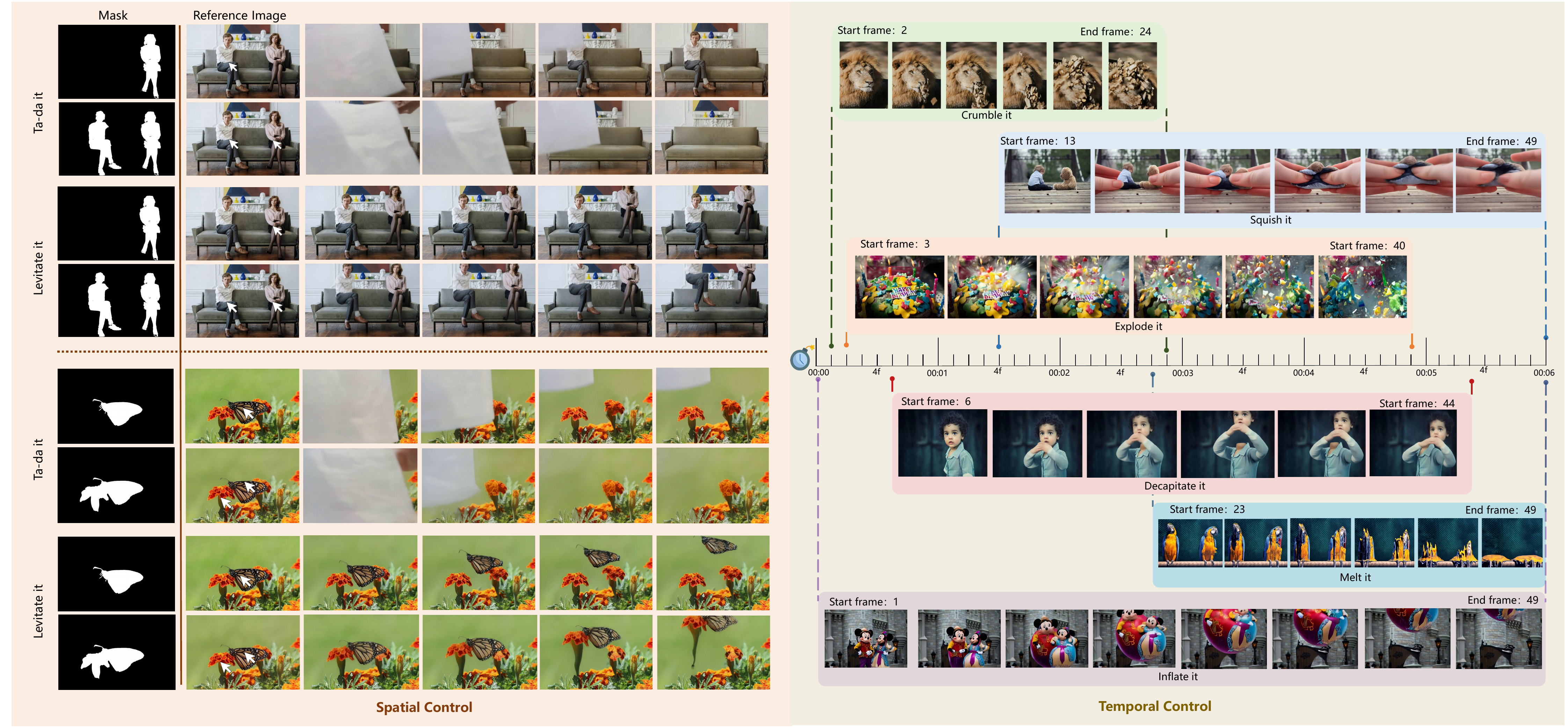}
        \caption{\textbf{VFX Creator} is an efficient framework based on a Video Diffusion Transformer, enabling spatial and temporal control for visual effect (VFX) video generation. With minimal training data, a plug-and-play mask control module allows precise instance-level manipulation, while the integration of tokenized start-end motion timestamps with text space provides fine-grained temporal control over the VFX rhythm.
}
\label{fig:illustration}
\end{teaserfigure}

\author{Xinyu Liu}
\affiliation{%
 \institution{Hong Kong University of Science and Technology}
 \country{China}}
\email{xliugd@connect.ust.hk}
\author{Ailing Zeng}
\affiliation{%
 \institution{Tencent AI Lab}
 \country{China}
}
\email{ailingzengzzz@gmail.com}
\author{Wei Xue}
\affiliation{%
 \institution{Hong Kong University of Science and Technology}
 \country{China}
}
\email{weixue@ust.hk}
\author{Harry Yang}
\affiliation{%
\institution{Hong Kong University of Science and Technology}
\country{China}}
\email{yangharry@ust.hk}
\author{Wenhan Luo}
\affiliation{%
 \institution{Hong Kong University of Science and Technology}
 \country{China}
}
\email{whluo@ust.hk}
\author{Qifeng Liu}
\affiliation{%
 \institution{Hong Kong University of Science and Technology}
 \country{China}}
\email{liuqifeng@ust.hk}
\author{Yike Guo}
\affiliation{%
 \institution{Hong Kong University of Science and Technology}
 \country{China}
}
\email{yikeguo@ust.hk}
\makeatletter
\def\@mkboth#1#2{\@mkbothcustom{#1}}
\def\@mkbothcustom#1{\markboth {\mytitle}{\mytitle}}
\makeatother

\title{VFX Creator: Animated Visual Effect Generation with Controllable Diffusion Transformer}
\begin{abstract}

Crafting magic and illusions stands as one of the most thrilling facets of filmmaking, with visual effects (VFX) serving as the powerhouse behind unforgettable cinematic experiences. While recent advances in generative artificial intelligence have catalyzed progress in generic image and video synthesis, the domain of controllable VFX generation remains comparatively underexplored. More importantly, fine-grained spatial-temporal controllability in VFX generation is critical, but challenging due to data scarcity, complex dynamics, and precision in spatial manipulation. In this work, we propose a novel paradigm for animated VFX generation as image animation, where dynamic effects are generated from user-friendly textual descriptions and static reference images. Our work makes two primary contributions: \textbf{i) Open-VFX}, the first high-quality VFX video dataset spanning 15 diverse effect categories, annotated with textual descriptions, instance segmentation masks for spatial conditioning, and start–end timestamps for temporal control; This dataset features a wide range of subjects for the reference images, including characters, animals, products, and scenes. \textbf{ii) VFX Creator}, a simple yet effective controllable VFX generation framework based on a Video Diffusion Transformer. The model incorporates a spatial and temporal controllable LoRA adapter, requiring minimal training videos. Specifically, a plug-and-play mask control module enables instance-level spatial manipulation, while tokenized start-end motion timestamps are embedded in the diffusion process accompanied by the text encoder, allowing precise temporal control over effect timing and pace. Extensive experiments on the Open-VFX test set with unseen reference images demonstrate the superiority of the proposed system to generate realistic and dynamic effects, achieving state-of-the-art performance and generalization ability in both spatial and temporal controllability. Furthermore, we introduce a specialized metric to evaluate the precision of temporal control. By bridging traditional VFX techniques with generative techniques, the proposed VFX Creator unlocks new possibilities for efficient, user-friendly, and high-quality video effect generation, making advanced VFX accessible to a broader audience.

\end{abstract}

\maketitle
\section{Introduction}

Visual effects (VFX) video generation is paramount in video production, particularly in cinema, gaming, and virtual reality, where it enhances visual impact and improves creative efficiency~\cite{adobe-guide}. Visual effects combine live action footage with generated imagery to create realistic environments, objects, animals, and creatures that would be dangerous, expensive, impractical, or impossible to capture on film. While early visual effects involved experimentation with film stock, modern techniques include animation, computer-generated imagery (CGI), and other post-production methods~\cite{chabanova2022vfx}. However, these approaches often involve high computational costs, long production cycles, and significant manual intervention. With the rapid development of diffusion models~\cite{blattmann2023stable,ho2020denoising}, visual effects generation is progressively transitioning from traditional techniques to generative models. 

Recent emerging models~\cite{lin2024open,polyak2024movie} have impressive video generation capabilities, showcasing strong temporal consistency and visually appealing effects. These advancements offer great potential for artists, enabling the creation of stunning videos with minimal input, such as images or text prompts. However, the field of VFX generation for video remains underexplored, with existing open-source models struggling to produce complex effects and effectively control motion generation from text prompts.

In contrast, closed-source products such as Pika~\cite{pika2023} and PixVerse~\cite{PixVerse} have proven their ability to generate a wide array of striking visual effects using diffusion-based generative models. These platforms can create effects such as realistic explosions, anti-gravity phenomena, and cinematic character transformations without manual modeling or lengthy production timelines. Despite their power, these proprietary platforms are limited by restricted access to their visual effects resources, which hinders broader development and exploration within this domain.

In this work, we propose a novel paradigm for animated VFX generation, such as image animation, where dynamic effects are generated from user-friendly textual descriptions and static reference images. We introduce two primary contributions to address the challenges and limitations of VFX video generation.
First, we present \textbf{Open-VFX}, the first high-quality generated VFX video dataset comprising 675 videos across 15 distinct effect categories, sourced from two commercial platforms: Pika and PixVerse. The dataset includes a diverse range of subjects—characters, animals, products, and scenes—with a minimum resolution of 700x1000 pixels. Additionally, it contains 245 static images from Pexels~\cite{pexels2024}, annotated with textual descriptions, frame-level masks, and normalized start-end timestamps. This wide-ranging dataset provides extensive reference material for the generation of VFX across various domains.
Second, we propose \textbf{VFX Creator}, a simple yet effective controllable VFX generation framework based on a Video Diffusion Transformer \cite{yang2024cogvideox}, as shown in Fig. \ref{fig:illustration}. The model incorporates a spatial and temporal controllable LoRA adapter, enabling high-quality video generation with minimal training data.
For spatial control, we integrate video mask sequences as conditions with the latent video noise, facilitating instance-level spatial manipulation. For temporal control, we tokenize the start-end motion timestamps and integrate them into the diffusion process alongside the text encoder, allowing for precise control over the timing and pacing of the effects.
Finally, we conduct comprehensive experiments on the Open-VFX dataset to demonstrate the model's generation capabilities. We also introduce a specialized evaluation metric to assess the precision of temporal control, showcasing the model's ability to generate dynamic, temporally consistent VFX.
Due to the data efficiency of VFX Creator, our framework can be easily adapted for fine-tuning across different categories of visual effects. This flexibility enables the rapid generation of a broad array of visual effects videos, significantly reducing the cost and time typically associated with traditional VFX production in the film industry.
In summary, our contributions are as follows:
\begin{enumerate}
\item We present the \textbf{Open-VFX}, the first high-quality VFX video dataset spanning 15 diverse effect categories, annotated with text prompt, instance segmentation masks for spatial conditioning, and start–end timestamps for temporal control.
\item We propose the \textbf{VFX Creator}, a simple yet effective controllable VFX generation framework based on
a Video Diffusion Transformer. The model incorporates a spatial and temporal controllable LoRA adapter, enabling precise manipulation.
\item We perform a comprehensive evaluation on the Open-VFX dataset, showcasing that the proposed system surpasses existing methods in generating visual effects. Additionally, we introduce a novel metric specifically designed to assess the precision of temporal control in the generated effects.
\end{enumerate}
\begin{figure*}[htbp]
    \centering
    \includegraphics[width=\textwidth]{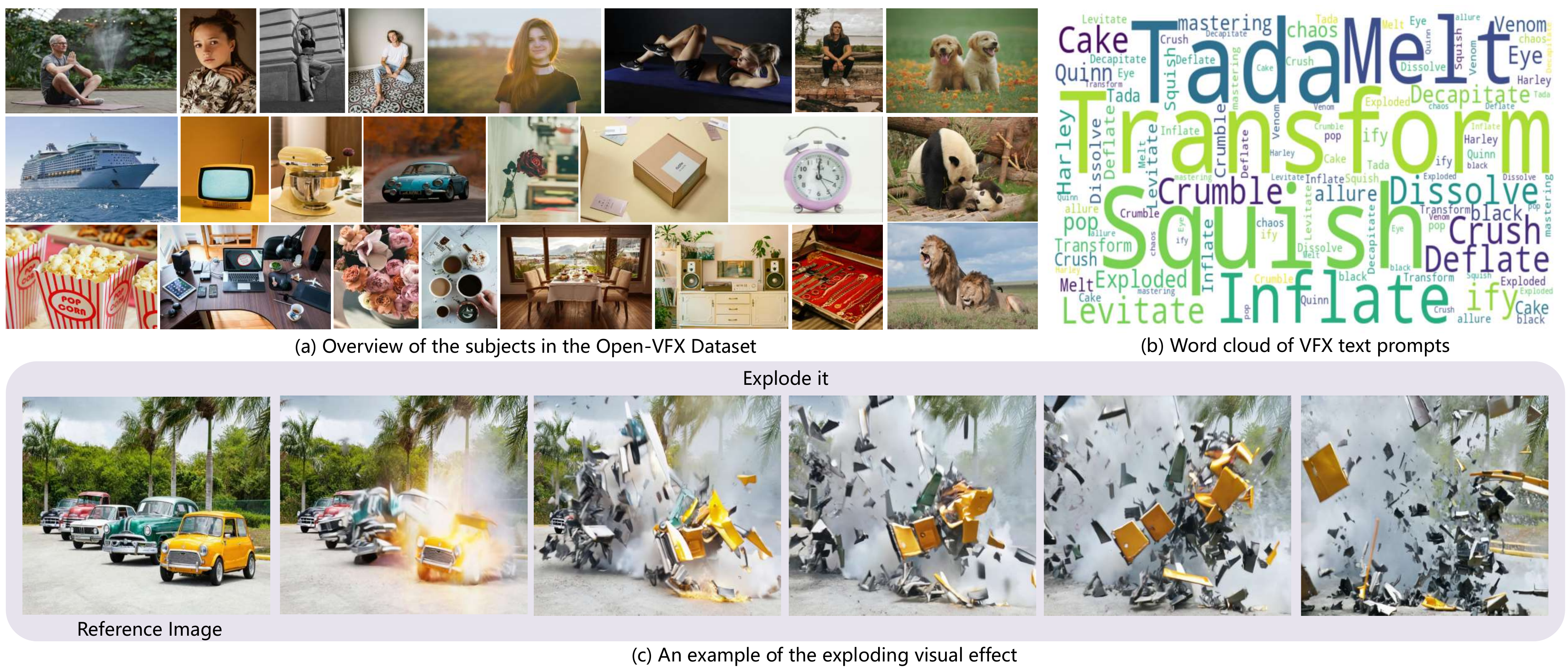}
    \caption{Overview of our proposed Open-VFX Dataset. (a) demonstrates diverse input inference images in the dataset, including humans, animals, objects, and various scenes across single and multiple components. (b) shows the text descriptions of the proposed 15 VFXs, and (c) presents an example (\emph{Explode it}) VFX.}
    \label{fig:Overview_Dataset}
\end{figure*}

\section{Related Work}

\subsection{General Video Generation}
The rapid advancement of video generative models is driven by diffusion models (DMs) \cite{sohl2015deep, rombach2022highresolutionimagesynthesislatent, ho2020denoising, song2020score, nichol2021glide}, which enable innovative approaches in video generation.
A key architecture is the Diffusion Transformer \cite{peebles2023scalable}, leveraging transformer designs to capture long-range dependencies, enhancing temporal consistency and dynamics, and multi-resolution synthesis~\cite{yang2024cogvideox, ma2024latte, shao2024human4dit, videoworldsimulators2024, chen2023videocrafter1, kuaishou2024keling, genmo2024mochi}.
For example, CogVideoX \cite{yang2024cogvideox} uses a 3D full attention mechanism for spatial and temporal coherence, while Hunyuan-DiT \cite{li2024hunyuanditpowerfulmultiresolutiondiffusion} introduces large pre-trained models for rich contextual detail.

Furthermore, controllable video generation has garnered considerable attention due to its promising applications in video editing and content creation. 
For instance, LAMP~\cite{wu2023lamp} focuses on transferring information from the first frame to subsequent frames, ensuring the consistency of the initial image throughout the video sequence; however, it is constrained by fixed motion patterns in a few-shot setting. 
Recent efforts have sought to enhance control over generative models by integrating additional neural networks into diffusion frameworks. 
ControlNet~\cite{zhang2023adding} directs image generation based on control signals by replicating specific layers from pre-trained models and connecting them with zero convolutions~\cite{wang2024disco}. However, the field of controllable visual effect video generation has yet to be explored.
\subsection{LoRA-Based Video Generation}
Recent advancements in fine-tuning methods~\cite{wu2024motionbooth,wu2024customcrafter,wang2024customvideo,guo2023animatediff,wang2024motionctrl,ouyang2024i2vedit} for video generation are often influenced by image customization techniques, especially LoRA~\cite{hu2021lora}. For example, Tune-A-Video~\cite{wu2023tune} extends a text-to-image model by introducing spatial-temporal attention and selectively training specific parts of the attention layers. Similarly, ~\cite{materzynska2023customizing} focus on fine-tuning only certain model components, with an emphasis on training earlier denoising steps to capture general motion rather than intricate appearance details. MotionDirector\cite{zhao2025motiondirector} proposes a dual-path LoRA architecture and an appearance-debiased temporal loss to decouple appearance and motion. Likewise, methods like~\cite{wei2024dreamvideo, zhang2023motioncrafter, ren2024customize} employ separate branches for appearance and motion. VMC~\cite{jeong2024vmc} adapts temporal attention layers by utilizing a motion distillation strategy, employing residual vectors between consecutive noisy latent frames to serve as motion references. Nevertheless, they have not yet investigated the combination of temporal and spatial controllability with LoRA to enhance the model's controllability.
\section{Dataset}
\subsection{Definition of Visual Effects}
Visual effects (VFX) can create realistic environments and characters that are difficult or impossible to capture during filming. For example, VFX involves compositing techniques to combine different visual elements into a single scene, often through green screen or digital background replacement. Additionally, digital effects such as explosions, smoke, and weather simulations help create dynamic and immersive environments. Motion capture technology is used to animate characters or creatures, while digital makeup effects transform actors into fantastical characters. In light of these various considerations, we have curated a set of \textbf{15} distinct visual effects, including \emph{Cake-ify, Crumble, Crush, Decapitate, Deflate, Dissolve, Explode, Eye-pop, Inflate, Levitate, Melt, Squish, Ta-da, Transformer into Venom, and Transformer into Harley Quinn}. As shown in Fig. \ref{fig:Overview_Dataset}, we demonstrate the overview of our Open-VFX Dataset, and these effects are designed to deliver a more immersive and visually compelling experience for the audience. Detailed descriptions of these effects can be found in the supplementary materials.
\begin{figure*}
    \centering
    \includegraphics[width=\textwidth]{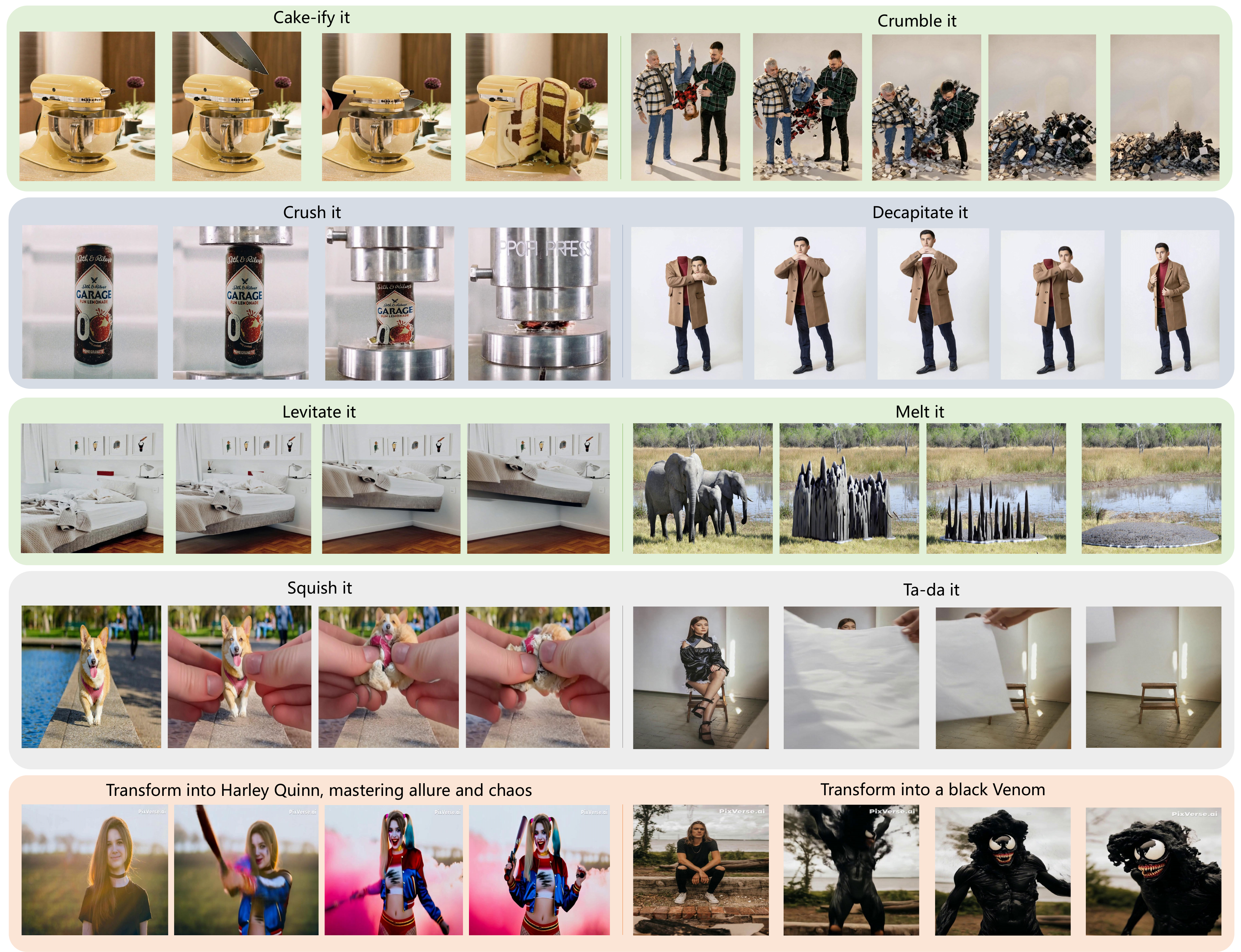}  %
    \caption{More examples of our Open-VFX dataset, including 10 VFXs and diverse reference images.}
    \label{fig:dataset}
\end{figure*}
\subsection{Source Videos}

The Open-VFX dataset comprises 675 high-quality VFX videos sourced from two commercial platforms, Pika 1.5~\cite{pika2023} and PixVerse 3.0~\cite{PixVerse}, with each video having an average duration of 5 seconds. Spanning 15 distinct categories of visual effects, the dataset also includes 245 reference images collected from the Pexels~\cite{pexels2024} community, featuring both single and multiple objects. As shown in Fig. \ref{fig:dataset}, all videos have a resolution of at least \(700 \times 1000\) pixels and are synthesized at 24 frames per second (fps).

\begin{figure}[htbp]
    \centering
    \includegraphics[width=0.5\textwidth]{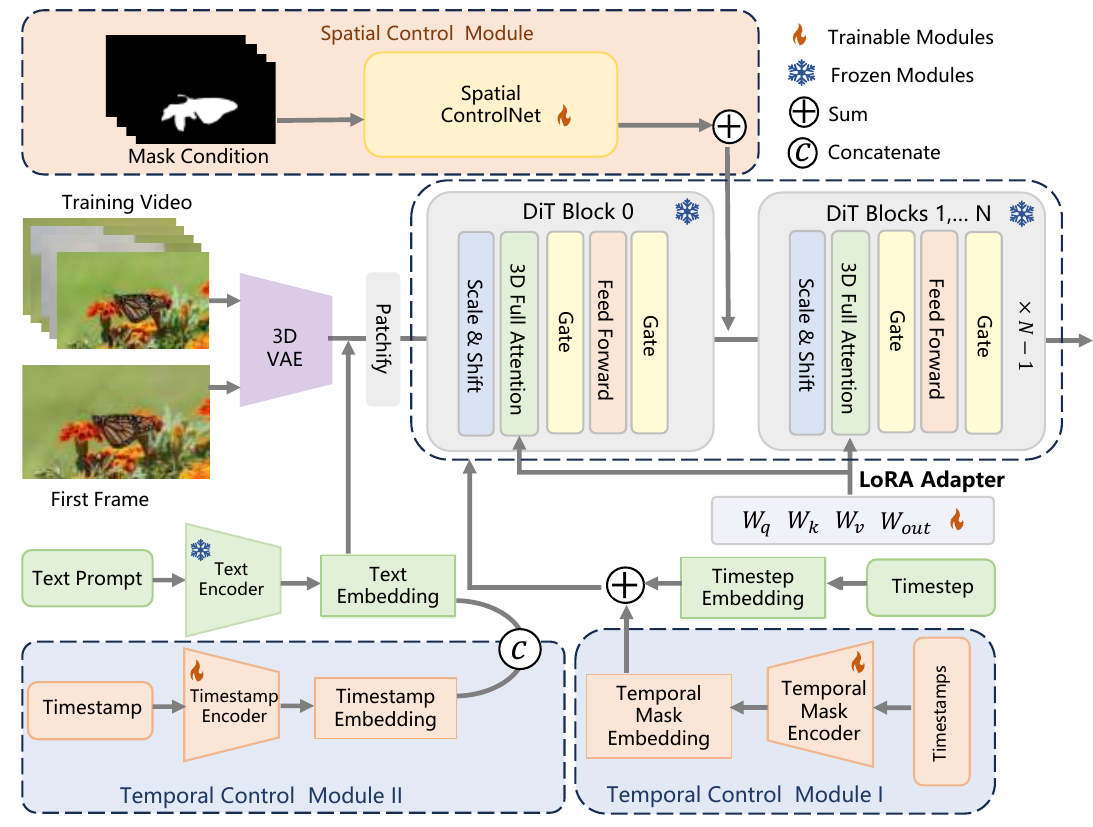} 
    \caption{\textbf{Pipeline of VFX Creator.} We introduce two novel modules: (a) Spatial Controlled LoRA Adapter. This module integrates a mask-conditioned ControlNet with LoRA, injecting mask sequences into the model to enable instance-level spatial manipulation. (b) Temporal Controlled LoRA Adapter. We explore two strategies for incorporating temporal control:  module I involves tokenizing start-end motion timestamps and embedding them into the diffusion process alongside the text space, while module II integrates temporal mask with timestep embeddings to guide the diffusion process.}
    \label{fig:main}
\end{figure}
\subsection{Data Annotation}
To accomplish the spatial-temporal controlled VFX video generation task, we adhered to the following process to acquire timestamp and mask annotations for our Open-VFX dataset:
\paragraph{Start-end motion timestamps.} To accurately capture the start and end timestamps of motion, traditional optical flow methods fall short in tracking the dynamic movement of visual effects due to their inherent limitations. Consequently, we employ Co-Tracker~\cite{karaev24cotracker3} for timestamp labeling. By monitoring the displacement of tracked points, motion is deemed to begin when the coordinates start to shift, and it concludes when the displacement ceases, providing precise temporal boundaries for the motion. 
\paragraph{Instance-level mask sequences.}  To enable the model to generate visual effects specifically tailored to the selected object, we utilize SAM2~\cite{ravi2024sam2segmentimages} to semantically annotate the motion of the chosen object, producing corresponding mask sequences. We generate several times to obtain diverse generated videos with different animated objects, and we annotate them via masks. During inference, SAM2 is employed to generate binary region masks based on the user-defined area, which serve as spatial conditions to guide precise control over the spatial manipulation of the object.

\section{Network}
\subsection{Preliminary}
We introduce the preliminary of CogVideoX architecture(Rombach et al., 2022), the baseline diffusion transformer Network used in our work, and Low-Rank Adaptation (LoRA) (Hu et al., 2021), which helps understand the spatial and temporal controllable LoRA adapter.
\subsubsection{Baseline Diffusion Transformer Network.}
VFX Creator builds upon the CogVideoX architecture~\cite{yang2024cogvideox} and leverages a causal 3D Variational Autoencoder (VAE)~\cite{kingma2013auto} for video compression, achieving temporal and spatial factors of 4 and 8, respectively. Latent variables are structured as sequential inputs, while textual information is encoded into embeddings using the T5 Encoder~\cite{raffel2020exploring}. These inputs are processed jointly within a stacked Expert Transformer network, which integrates Adaptive Layer Normalization for better alignment and 3D Rotary Positional Embeddings (RoPE)~\cite{narvekar2011no} to enhance the model's ability to capture temporal dynamics and long-range dependencies in video frames.
\subsubsection{Low-rank Adaptation(LoRA)}
LoRA~\cite{hu2021lora} targets the residual component of the model, denoted as $\Delta W$, which is added to the original weight matrix, yielding the updated weights as: $W^{\prime}=W+\Delta W$. In this formulation, $\Delta W$ is expressed as the product of two low-rank matrices: $\Delta W=AB^{T}$, where $A\in\mathbb{R}^{n\times d}$, $B\in\mathbb{R}^{m\times d}$, and $d<n$, $d<m$. By focusing on the smaller low-rank matrices $A$ and $B$, rather than the full-weight matrix $W$, LoRA effectively reduces both computational and memory costs during the training process.
\subsection{VFX Creator}
As shown in Fig. \ref{fig:main}, VFX Creator is a controllable VFX generation framework based on a Video Diffusion Transformer. The model incorporates a spatial and temporal controllable LoRA adapter, requiring minimal training videos. Specifically, the mask control module enables instance-level spatial manipulation, while tokenized start-end motion timestamps are embedded in the diffusion process accompanied by the text encoder, allowing precise temporal control over effect timing and pace.

\subsubsection{\textbf{Temporal Controlled LoRA Adapter.}}

To enable rhythm-controlled visual effect generation, we first employ temporal video augmentation by utilizing the start and end timestamps of the effect's motion to maximize data utilization. Specifically, the moving video clip is subject to a random shift, with the start and end timestamps constrained within the ranges of $[0, T-D]$ and $[D, T]$, where $T$ represents the maximum frame count in the training video, and $D$ denotes the duration of the visual effect's motion.

In this section, we explore two strategies for integrating timestamp control signals and conduct experiments on three different effects to analyze their effectiveness and accuracy.
\paragraph{\textbf{Strategy I: Integrate Temporal Mask with Timestep.}}
Current temporal condition representation typically takes two forms: digitized start-end timestamps and temporal masks. The former normalizes timestamps directly, while the latter applies a temporal mask to the frame sequence, designating moving frames as 1 and static frames as 0. 
Existing methods for temporal condition injection can be categorized into two strategies: one integrates temporal mask with Timestep embeddings and injects them into the diffusion blocks; the other utilizes a timestamp encoder to interact with the text space and computes cross-attention with the noisy latent representation. In the following, we first focus on the former approach.

We project the temporal mask of motion in videos into a timestep embedding and add it to each frame to ensure uniform application of the motion's timing and pace to every frame. 
As illustrated in Module I of Fig.~\ref{fig:main}, given the normalized timestamp, we introduce a timestamp encoder network that projects the temporal masks into the timestep embedding space, which is then incorporated into the DiT blocks~\cite{peebles2023scalable}.
\paragraph{\textbf{Strategy II: Integrate Timestamps with Text Space.}}
Inspired by ~\cite{fang2024camera}, we first map the start and end timestamps of the visual effects to the prompt space, converting them into timestamp tokens. These tokens are then concatenated with the original text prompt tokens. By leveraging the cross-attention mechanism in the DiT block, we generate VFX videos conditioned on the temporal control signals. As shown in Module II of Fig.~\ref{fig:main}, given the normalized timestamps $y_{\mathrm{timestamp}}$, the text prompt $y_{\mathrm{text}}$, and text-domain specific encoder $\tau_{\theta}$, we introduce a timestamps encoder network $\tau_{\phi}$, that projects timestamps $y_{\mathrm{timestamp}}$ to the intermediate text representation space $\mathbb{R}^{M\times d_{\tau}}$ such that
\begin{equation}
    \tau_\phi(y_{timestamp})\in\mathbb{R}^{M\times d_\tau}
\end{equation}
We concatenate $\tau_{\phi}(y_{\mathrm{timestamp}})$ and$\tau_\theta(y_{\mathrm{text}})$ and input the cross-attention layer Attention $(Q,K,V)={\mathrm{softmax}}\left({\frac{QK^{T}}{\sqrt{d}}}\right)\cdot V$,with
\begin{equation}
    \begin{aligned}&\mathrm{Q}=W_{Q}^{(i)}\cdot\varphi_{i}(z_{t}),\\&\mathrm{K}=W_{K}^{(i)}\cdot(\tau_{\phi}(y_{\mathrm{timestamp}})\oplus\tau_{\theta}(y_{\mathrm{text}})),\\&\mathrm{V}=W_{V}^{(i)}\cdot(\tau_{\phi}(y_{\mathrm{timestamp}})\oplus\tau_{\theta}(y_{\mathrm{text}})).\end{aligned}
\end{equation}
where $z_{t}$ denotes the latent representation  $z$ at the  $t$-th diffusion time step. $\varphi_{i}(z_{t})\in\mathbb{R}^{N\times d_{\epsilon}^{i}}$ denotes a flattened intermediate representation of the Transformer implementing $\epsilon_{\theta}$. $W_{V}^{(i)}\in\mathbb{R}^{d\times d_{\epsilon}^{i}}$ and $W_{Q}^{(i)}\in\mathbb{R}^{d\times d_{\tau}}$ and $W_K^{(i)}\in\mathbb{R}^{d\times d_\tau}$ are learnable projection matrices. $\oplus$ is the operator for tensor concatenation.

\subsubsection{\textbf{Spatial Controllable LoRA Adapter.}}
Currently, we commonly employ three methods to incorporate spatial conditions into the diffusion model:
i) concatenating the reference image with the first frame mask before inputting them into the diffusion model~\cite{ma2024follow}; ii) combining the latent mask and noisy latent along the channel dimension for spatial control~\cite{lei2024animateanything}; iii) introducing a spatial ControlNet~\cite{zhang2023adding} to extract a mask sequence that guides the generation process.

Our experimental observations reveal that the first two methods do not successfully facilitate spatial condition-based video generation tasks. While effective in U-Net-based diffusion models, they are not suitable injection techniques for transformer-based diffusion models.
Therefore, to enable instance-level spatial manipulation, we introduce a plug-and-
play mask control module, leveraging the mask guidance to precisely control the desiring instance.

During training, we utilize SAM2 to obtain the mask sequences of the moving instances. We retain the mask sequences preceding the start timestamp, while padding the remaining frames with zeros, and combine them to form the spatial condition. These spatial conditions are then injected into a learnable spatial ControlNet~\cite{zhang2023adding} for fine-tuning. As shown in Fig.~\ref{fig:main}, we extract mask conditions and integrate them into the main network. This branch shares trainable parameters initialized as a copy of the original half branch and operates in parallel, using a zero convolution as a bridge to integrate the conditional controls. Specifically:
\begin{equation}
    \boldsymbol{y}_c=\mathcal{F}_m(\boldsymbol{x})+\mathcal{Z}(\mathcal{F}_{cn}(\boldsymbol{x},\boldsymbol{c};\Theta_{cn});\Theta_z),
\end{equation}
where $\mathcal{F}(\cdot;\Theta)$ denotes a neural model with learnable parameters $\Theta,\mathcal{Z}(\cdot;\Theta_{z})$ indicates the zero convolution layer, and $x,y_{c}\in\mathbb{R}^{h\times w\times c}$ and $c$ are the 2D feature maps and conditional controls, respectively. This trainable spatial ControlNet branch is connected to the partly frozen main branch with a zero-initialized convolution layer, ensuring the integration of spatial conditions while minimizing interference with the base model.

\section{EXPERIMENT}
To assess the effectiveness of our Open-VFX dataset and VFX Creator model, we conduct experiments across all of the visual effects, demonstrating VFX Creator's versatility in generating controllable VFX videos. To assess the performance of VFX video generation, we compare VFX Creator with the following state-of-the-art methods: (i) DynamiCrafter~\cite{xing2025dynamicrafter}, (ii) CogVideoX~\cite{yang2024cogvideox}, (iii) LTX-Video~\cite{hacohen2024ltx}, and (iv) the pseudo ground-truth Pika or PixVerse. All of these approaches allow text prompts and reference images as input. Furthermore, we select several visual effects to conduct a focused evaluation of both temporal and spatial control accuracy. 
\begin{figure*}[htbp]
    \centering
    \includegraphics[width=\textwidth]{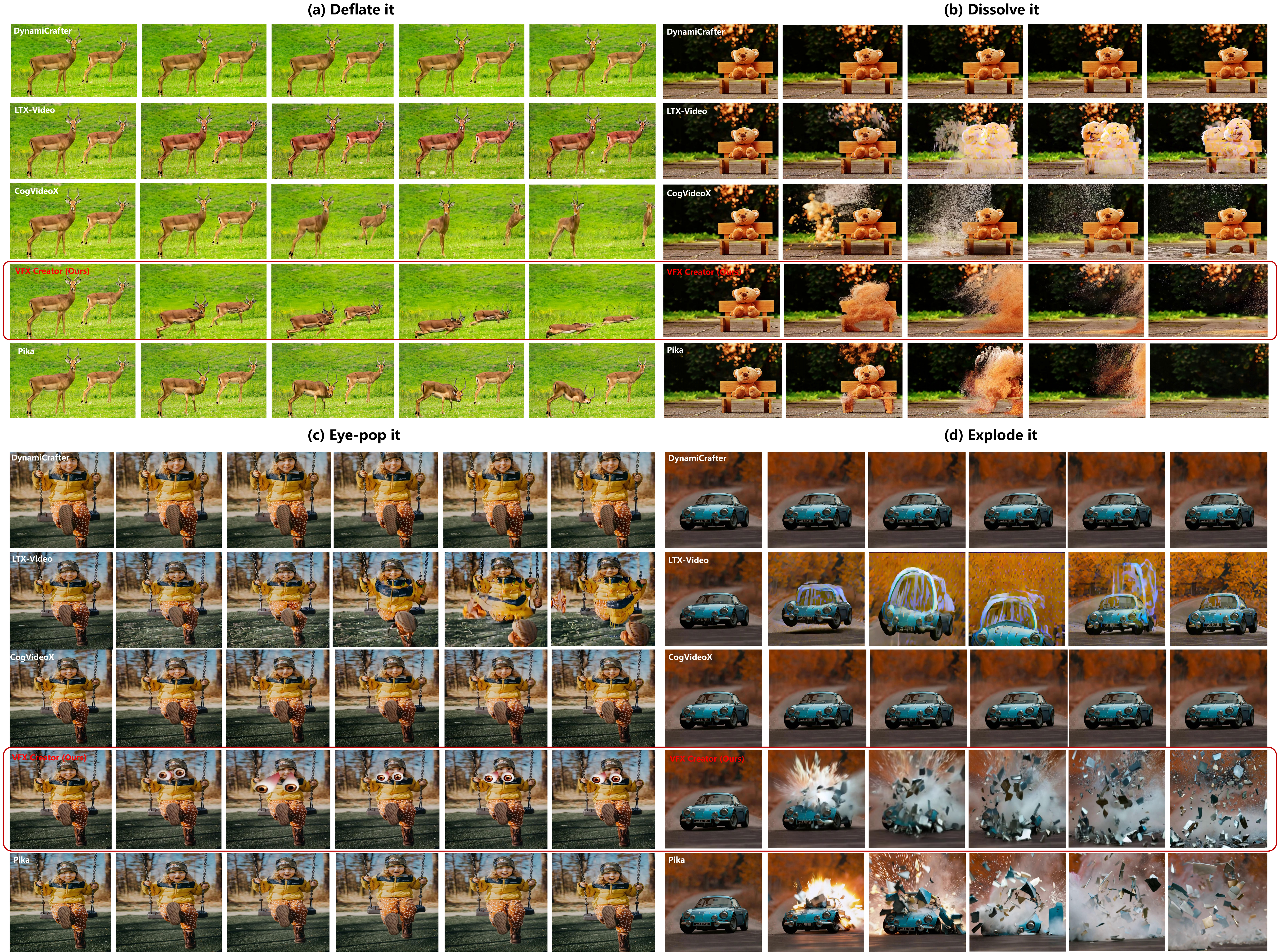}
    \caption{Qualitative comparisons of VFX video generation on two different visual effects between our method, DynamiCrafter, LTX-Video, CogVideoX-5B, and Pika.}
    \label{fig:compare}
\end{figure*}

\subsection{Implementation Details}
During the training phase, we incorporate low-rank matrices with a rank of 128 into the 3D Transformer module of the baseline network. We randomly sample 49 frames with a resolution of 480 × 720. We employ the AdamW~\cite{loshchilov2017decoupled} optimizer with a constant learning rate of 1e-4 for training all models. All experiments are conducted on the NVIDIA H800 GPU.
We froze the gradients of most weights in the original base network and trained for 3000 steps with a learning rate of 1e-4, implementing both learning rate warm-up and decay mechanisms.
Our VFX dataset is partitioned into training and testing sets in a 9:1 ratio, both containing 15 distinct types of visual effects. The training set includes an average of 40 videos per effect, while the testing set contains 5 videos per effect. 

\subsection{Evaluation Metrics}
In this experiment, we adopt three metrics following prior works: FID-VID~\cite{unterthiner2018towards}, FVD~\cite{balaji2019conditional} and Dynamic Degree~\cite{huang2024vbench} to evaluate the general quality and degree of dynamics of the synthesized videos. 
As shown in Table 2, we aim to focus on the generative performance of the motion in the video.  
We design three metrics to evaluate the accuracy of temporal control: frame-level errors $\mathcal{E}_{\text{f}} $ and second-level errors $\mathcal{E}_{\text{s}} $, as well as Temporal Intersection over Union ($T_{\text{IoU}}$). Specifically, temporal error quantifies the difference between the start and end timestamps of the predicted and ground truth segments. The frame-level temporal error and second-level temporal error are related through the frames per second (FPS):
\begin{equation}
    \mathcal{E}_{\text{f}}   =\frac{1}{N}\sum_{i=1}^{N}\left(\left|\hat{t}_{\mathrm{start},i}-t_{\mathrm{start},i}\right|+\left|\hat{t}_{\mathrm{end},i}-t_{\mathrm{end},i}\right|\right)
\end{equation}
\begin{equation}
    \mathcal{E}_{\text{s}}   = \frac{1}{N} \sum_{i=1}^{N} \left( \left| \frac{\hat{t}_{\mathrm{start},i}}{\mathrm{FPS}} - \frac{t_{\mathrm{start},i}}{\mathrm{FPS}} \right| + \left| \frac{\hat{t}_{\mathrm{end},i}}{\mathrm{FPS}} - \frac{t_{\mathrm{end},i}}{\mathrm{FPS}} \right| \right)
\end{equation}
where \( N \) is the number of video segments, and \( t_{\mathrm{start},i} \) and \( t_{\mathrm{end},i} \) are the normalized start and end timestamps of the \( i \)-th video clip.

To evaluate the accuracy of the predicted timestamps, we randomly sample five pairs of start and end ground truth timestamps for a reference sample. The start ground truth is constrained within the range of $[0, 2/T]$, while the end ground truth falls within the range of $[2/T, T]$, where $T$ represents the total number of frames.

\subsection{Quantitative Comparison Results}
For quantitative evaluation, we conducted experiments on 15 types of visual effects from our dataset and compared VFX Creator with three state-of-the-art open-source methods, with detailed results shown in Table \ref{tab:overall_compare}. VFX Creator outperforms the other methods across all metrics, particularly in generating visual effects with large motion patterns, indicating superior video quality and more accurate motion generation.
As seen in Table \ref{tab:overall_compare}, DynamiCrafter~\cite{xing2025dynamicrafter} and CogVideoX~\cite{yang2024cogvideox} are less responsive to visual effect prompts, producing videos with minimal or no motion, as reflected by their lower dynamic degrees. While LTX-Video~\cite{hacohen2024ltx} exhibits a higher dynamic degree, this is due to the generation of large, incorrect motions, which does not correspond to the semantic accuracy of the motion. These results align with the limitations observed in the quantitative evaluation.
To further validate these findings, we conducted extensive user studies to assess the correspondence accuracy between generated visual effects and text prompts. VFX Creator consistently outperforms other state-of-the-art methods, confirming its exceptional ability to generate high-quality, semantically accurate visual effects, even when dealing with large and complex abstract motions.

\begin{table}[h]
\centering
\caption{Quantitative comparisons of VFX video generation for 15 visual effects in our dataset.
}
\scalebox{0.8}{
\renewcommand{\arraystretch}{1.1} 
\begin{tabular}{c|cccc}
\toprule
Method & FID-VID $\downarrow$ & FVD $\downarrow$ & Dynamic Degree $\uparrow$\\
\midrule
DynamiCrafter & 119.78 & 1515.28 & 0.27 \\
LTX-Video & 82.93 & 1563.73 & 0.51  \\
CogVideoX & 90.82 & 1624.91 & 0.14  \\
VFX Creator (Ours) & \textbf{29.92} & \textbf{752.95} & \textbf{0.63}  \\
\bottomrule
\end{tabular}}
\renewcommand{\arraystretch}{1} 
\label{tab:overall_compare}
\end{table}

\subsection{Qualitative Comparison Results}

\begin{figure*}
    \centering
    \includegraphics[width=\textwidth]{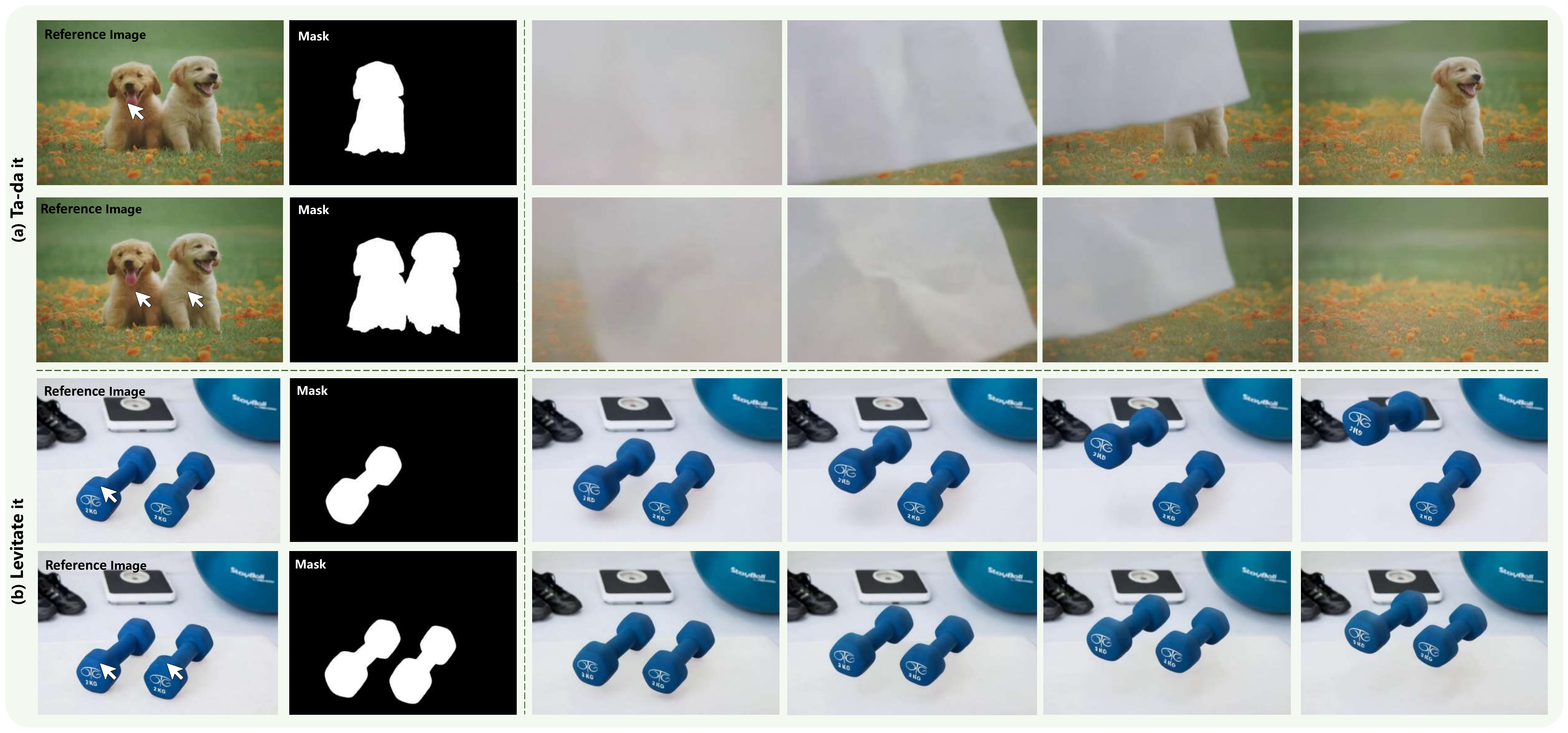}  %
    \caption{Qualitative results of spatial controllable VFX video generation of our method on two different visual effects. Users can precisely specify the animated instance by clicking points or dropping boxes to obtain the mask.}
    \label{fig:spatial}
\end{figure*}

We present qualitative comparison results of VFX Creator with four representative models on three effects: "Deflate it," "Dissolve it," and "Eye-pop it," as shown in Fig.~\ref{fig:compare}. For comparison, results from Pika are used as the benchmark. We found that VFX Creator consistently generates more reliable videos, even when Pika struggles to produce satisfactory outputs. For example, Pika fails to generate correct effects in "Defalte it" and "Eye-pop it", while VFX Creator successfully generates the abstract effect video. Pika's failure may stem from the substantial increase in complexity associated with accurately locating and generating the eye in smaller, more distant contexts. In contrast, DynamicCrafter exhibits significant challenges in generating visual effect motions that correspond to text prompts. LTX-Video and CogVideoX-5B, on the other hand, generate either small-scale motions or large, erroneous ones, which leads to a lack of temporal consistency and semantic accuracy. These limitations are not present in VFX Creator, which maintains better alignment with the intended motion and effect characteristics. This ensures improved temporal consistency and visual fidelity.
These results demonstrate that VFX Creator outperforms other models, delivering visually accurate and temporally consistent visual effects, even in cases where ground truth struggles with reliability. 
\subsection{Ablation Study}

The proposed VFX Creator is pivotal for generating high-quality controllable VFX videos. To assess the contributions of the spatial and temporal control modules, we conduct a series of ablation studies. \textbf{First}, we compare the results of two temporal control injection strategies. The impact of start and end timestamp guidance on animation results is presented in Table~\ref{tab:model_design_ablation}. Our observation reveals that integrating timestamps with the text space yields superior results. This approach integrates timestamps with textual prompts to perform a cross-attention mechanism with video latent representations, enhancing the precise alignment of visual effects with specified temporal cues. Additionally, the transformer-based diffusion model is more effective in handling implicit representation injections compared to explicit conditions. \textbf{Additionally}, we present the quantitative results of integrating the spatial control module. As demonstrated in Fig.~\ref{fig:spatial}, VFX Creator effectively achieves accurate object manipulation, generating photorealistic videos with strong consistency between the visual effects and user interactions. \textbf{Lastly}, we analyze the impact of different sample sizes during training by comparing the model's performance across varying shot numbers: 1-shot, 10-shot, and 40-shot. As shown in Table~\ref{tab:model_design_ablation}, we observe that the number of shots plays a significant role in the model's performance. The results indicate that increasing the number of shots generally improves performance, with the 10-shot configuration often yielding balanced results. This highlights the model's data efficiency, demonstrating its ability to learn abstract and complex visual effect motions without the need for large amounts of data.

\begin{table}[h]
\centering
\caption{Ablation results of two temporal control integration strategies.}
\scalebox{0.85}{
\renewcommand{\arraystretch}{1.2} 
\begin{tabular}{c|ccc|ccc}
\toprule
\multirow{2}{*}{Visual Effect} & \multicolumn{3}{c|}{Temporal Strategy I} & \multicolumn{3}{c}{Temporal Strategy II} \\
                                                          & $T_{\text{IoU}} \uparrow$ & $\mathcal{E}_{\text{f}} \downarrow$ & $\mathcal{E}_{\text{s}} \downarrow$ & $T_{\text{IoU}} \uparrow$ & $\mathcal{E}_{\text{f}} \downarrow$ & $\mathcal{E}_{\text{s}} \downarrow$ \\
\midrule
Ta-da it       & 0.69 & 12.52 & 1.56& 0.85 & 5.04 & 0.63 \\
Explode it      & 0.68 & 11.30 & 1.49& 0.88 & 3.76 & 0.47  \\
Levitate it    & 0.69 & 12.88 & 1.61& 0.80 & 5.36 & 0.67  \\
\midrule
Average       & 0.69 & 12.23 & 1.56& \textbf{0.84} & \textbf{4.72} & \textbf{0.59}\\
\bottomrule
\end{tabular}}
\renewcommand{\arraystretch}{1} 
\label{tab:timestamps_two_strategy}
\end{table}

\begin{table}[h]
\centering
\caption{Ablation results of different sample sizes during training across varying shot numbers.}
\renewcommand{\arraystretch}{0.8} 
\scalebox{0.85}{
\begin{tabular}{c|c|ccc}
\toprule
 Effect & Shots& FID-VID $\downarrow$ & FVD  $\downarrow$ & Dynamic Degree  $\uparrow$\\
\midrule
\multirow{3}{*}{Ta-da it}&1-shot & 52.31& 1432.40 & 0.6   \\
&10-shot & \textbf{47.91} & 2861.18& 1.0  \\
&40-shot & 54.73 & \textbf{726.83}&  \textbf{1.0}   \\
\midrule
\multirow{3}{*}{Explode it}&1-shot & 96.48&2667.72 & 1.0  \\
&10-shot &  57.71 &2829.00& 1.0 \\
&40-shot & \textbf{50.97} & \textbf{2394.20}& \textbf{1.0}  \\
\midrule
\multirow{3}{*}{Squish it}&1-shot &140.42 & 3297.11& 1.0   \\
&10-shot &  44.62 &1409.98& 1.0  \\
&40-shot & \textbf{44.35} &  \textbf{1644.69}& \textbf{1.0}   \\
\bottomrule
\end{tabular}}
\renewcommand{\arraystretch}{1} 
\label{tab:model_design_ablation}
\end{table}

\subsection{User Study}
\begin{figure}[htbp]
    \centering
    \includegraphics[width=0.45\textwidth]{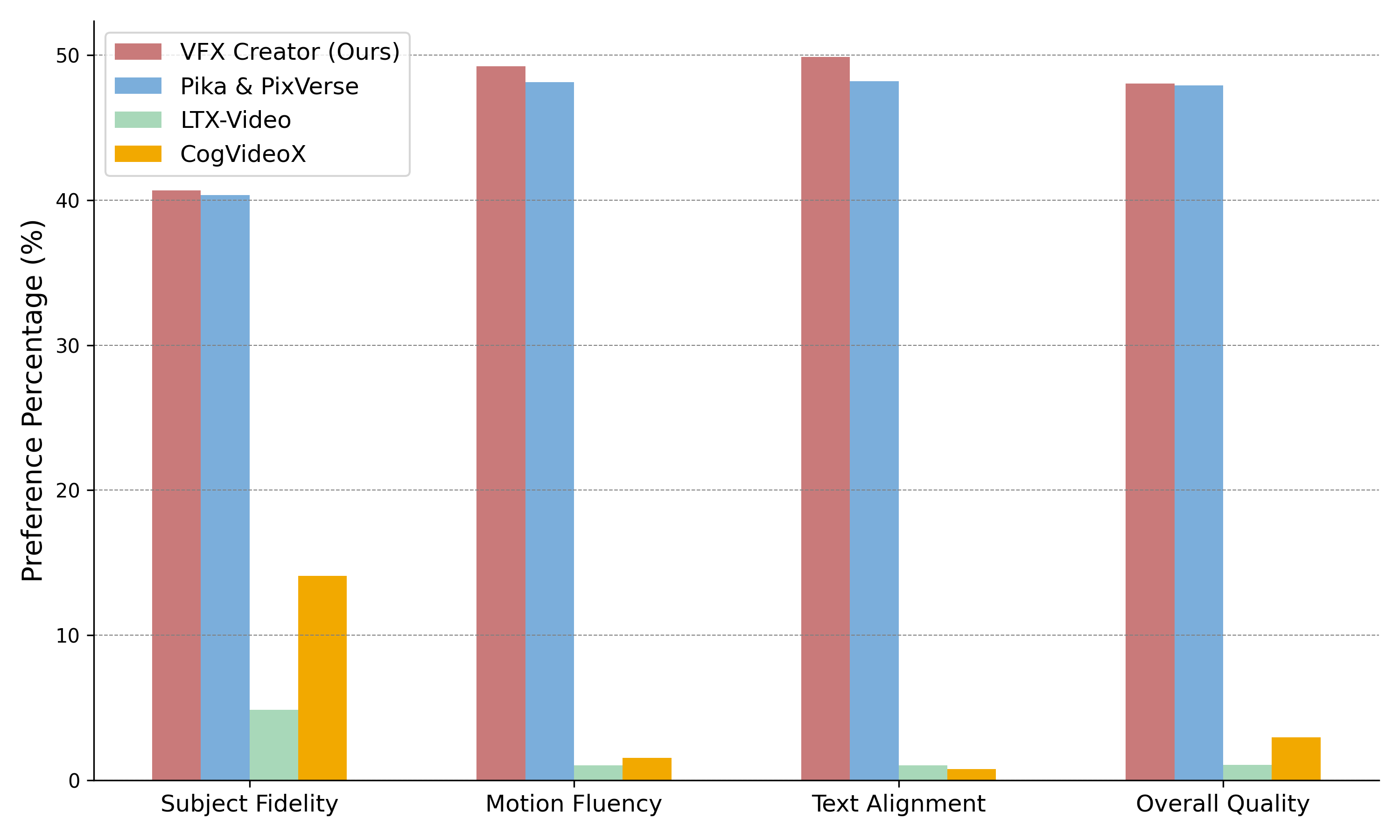} 
    \caption{\textbf{User Study}. Our VF Creator demonstrates superior human preference compared to other methods}
    \label{fig:user_study}
\end{figure}

To further validate the effectiveness of our method, we conducted a human evaluation, comparing our approach with four existing approaches, without using additional data for guidance. We invited 20 users to assess 30 sets of generated comparing results. We evaluated the quality of the generated videos across four dimensions: \emph{Text Alignment, Subject Fidelity, Motion Fluency, and Overall Quality}. Text Alignment measures how well the generated video aligns with the text prompt; Subject Fidelity evaluates how closely the generated object matches the reference image; Motion Fluency assesses the smoothness and quality of the motions in the generated video; and Overall Quality reflects whether the overall quality of the generated video meets user expectations. As shown in Fig.~\ref{fig:user_study}, both our method, VFX Creator, and Pika\& PixVerse achieved superior user preference across all metrics, with VFX Creator slightly outperforming Pika\& PixVerse, demonstrating the effectiveness of our approach.
\section{Limitation}
Despite the introduction of our pioneering visual effects video dataset, Open-VFX, alongside the development of VFX Creator for spatial-temporal controllable effect generation, several limitations remain. Firstly, while we have curated 15 types of effects and designed the system for data efficiency to facilitate easy VFX personalization, there is still considerable scope for expanding the dataset in both breadth and depth to address the diverse visual effects requirements across various scenes. Furthermore, we explored training a unified model for visual effect generation. However, experimental results indicate that the quality of unified training falls short compared to category-specific training, as direct unification tends to confuse multiple visual effects. In the future, implementing strategies such as Mixture of Experts (MOE)~\cite{chen2022understandingmixtureexpertsdeep} may improve the performance of mixed training approaches.
\section{Conclusion}
In conclusion, this work presents significant advancements in the field of controllable visual effects (VFX) generation, addressing critical challenges associated with fine-grained spatial and temporal manipulation. First, we propose the Open-VFX dataset establishes a foundational resource for future explorations in VFX generation, offering a diverse array of effect categories and detailed annotations that enhance the training of VFX generation models. Furthermore, we introduce VFX Creator, a simple yet effective controllable VFX generation framework based on
a Video Diffusion Transformer. Specifically, we leverage minimal training videos and enabling fine-grained spatial and temporal control, our system bridges the gap between traditional VFX techniques and modern generative models. The extensive experiments demonstrate its ability to produce realistic, dynamic effects with state-of-the-art performance in both spatial and temporal controllability. With the innovative integration of instance-level spatial manipulation and precise temporal control, VFX Creator paves the way for efficient and user-friendly VFX generation, making advanced visual effects more accessible to a broader audience and expanding creative possibilities in filmmaking.

\bibliographystyle{ACM-Reference-Format}
\bibliography{sample-bibliography}

\newpage
\appendix
\section{Definations of visual effects}
The Open VFX dataset encompasses 15 distinct categories of visual effects, featuring a wide array of subjects, including characters, animals, products, and scenes. As illustrated in Table~\ref{tab:define}, we offer a comprehensive explanation of the specific meanings associated with each visual effect, facilitating a deeper understanding for users.
\begin{table}[H]
\centering
\caption{Visual Effect Types and Corresponding Definitions in the Open VFX Dataset.}
\label{tab:dataset_define} 
\begin{tabular}{p{2cm}|p{5.8cm}} 
\hline
Types of VFX & Definition \\
\midrule
\multirow{2}{*}{Cake-ify it} & Transform the subject into hyper-realistic prop cakes. \\
\midrule
Crumble it&  Break apart the subjects into fragments. \\
\midrule
Crush it & Apply a hydraulic press to flatten the subject. \\
\midrule
Decapitate it & Simulate the decapitation of subjects. \\
\midrule
\multirow{2}{*}{Deflate it} & Similar to a balloon losing air, cause subjects to shrink and flatten. \\
\midrule
\multirow{2}{*}{Dissolve it} & Cause the object to disintegrate into nothingness.\\
\midrule
Explode it & Burst the subject into fragments.\\
\midrule
Eye-pop it & Make the eyes of subjects bulge or pop out.\\
\midrule
Inflate it & Puff up the still subject like a balloon.\\
\midrule
\multirow{2}{*}{Levitate it} &  Make static objects or subjects appear to float or hover.\\
\midrule
Melt it &Turn objects into fluid, gooey forms. \\
\midrule
\multirow{2}{*}{Squish it} & Compress the subject as though under immense pressure.\\
\midrule
\multirow{2}{*}{Ta-da it} & With a flourish, subjects disappear behind a cloth.\\
\midrule
Transform into a black Venom &Characterize the static subject, transforming it into a black Venom.\\
\midrule
Transform into Harley Quinn& Characterize the static subject, transform it into Harley Quinn.\\
\hline
\end{tabular}
\label{tab:define}
\end{table}
\section{Quantitative results of mixed training}
As a data-efficient system capable of achieving visual effect personalization, we attempted to explore a unified VFX generation model. Specifically, we tuned 15 effects using 600 effect videos in a combined manner. We then evaluated the results of the unified model, calculating their FID-VID~\cite{unterthiner2018towards}, ~\cite{balaji2019conditional}, and Dynamic Degree~\cite{huang2024vbench}. Additionally, we compared the results of the 15 effects from both category-specific training and unified and mixed training. As shown in Table ~\ref{tab:control_results}, experimental results indicate that the quality of unified training falls short compared to category-specific training, as direct unification tends to confuse multiple visual effects. Most of the effect results indicate that mixed training leads to a decline in overall video quality, although about one-third of the effects exhibit slight improvements after mixed training, such as "Ta-da it." In the future, implementing strategies such as Mixture of Experts (MOE)~\cite{chen2022understandingmixtureexpertsdeep} may improve the performance of mixed training approaches.

\begin{table*}
\centering
\caption{Comparison results of mix and single training for visual effect generation.}
\renewcommand{\arraystretch}{1.1} 
\begin{footnotesize} 
\begin{tabular}{@{}c|c|ccccccccccccccc@{}} 
\toprule
\rotatebox{90}{Metric} &\rotatebox{90}{Method}&\rotatebox{90}{Cake-ify}  & \rotatebox{90}{Crumble} & \rotatebox{90}{Crush} & \rotatebox{90}{Decapitate} & \rotatebox{90}{Deflate} & \rotatebox{90}{Dissolve}& \rotatebox{90}{Explode}& \rotatebox{90}{Eye-pop} & \rotatebox{90}{harley} & \rotatebox{90}{Inflate} & \rotatebox{90}{Levitate} & \rotatebox{90}{Melt}& \rotatebox{90}{Squish} & \rotatebox{90}{Ta-da}& \rotatebox{90}{Venom}\\
\midrule
\multirow{2}{*}{FID-VID$\downarrow$}   & Single & \textbf{54.48}        & 65.11    &46.71     & \textbf{43.76}          & \textbf{103.90}      & \textbf{76.14} &\textbf{50.97}&\textbf{34.87}&\textbf{94.62}&86.14&\textbf{35.12}&\textbf{63.37}&\textbf{44.35}&54.73&117.90     \\
                           & Mix& 67.22        & \textbf{65.06}    & \textbf{44.52}     & 44.52          & 111.28      & 87.00 &84.19&54.43&117.34&\textbf{77.35}&68.32&70.38 &52.36&\textbf{34.65}&\textbf{108.99}   \\
\midrule
\multirow{2}{*}{FVD$\downarrow$}        & Single&\textbf{1140}        & \textbf{1690}    & \textbf{1000}     & 1263          & \textbf{2034}      & \textbf{1463}&\textbf{2394}&\textbf{1547}&\textbf{3566}&\textbf{1946}&\textbf{665}&\textbf{1794}&1644 &\textbf{726} &\textbf{3668}   \\
                           & Mix&1503        & 1738    &1015     & \textbf{1054}          & 2133      & 1794&2612&1641&3811&2184&1018&2774&\textbf{1543} &979 &3911    \\
\midrule
\multirow{2}{*}{Dynamic Degree$\uparrow$} & Single& 0.8     & 0.8         & 0.0          & 0.6      & 0.0 & 0.8& 1.0& 0.0& 1.0& 0.8& 0.0& 0.6 & 1.0& 1.0 & 1.0   \\
                               & Mix& 0.8     & 0.8         & 0.0         &  0.6      &  0.0 & 0.8& 1.0& 0.0& 1.0& 0.8& 0.0 & 0.6 & 1.0 & 1.0 & 1.0 \\
\bottomrule
\end{tabular}
\end{footnotesize}
\renewcommand{\arraystretch}{1} 
\label{tab:control_results}
\end{table*}

\section{More Controlled Qualitative results }
\begin{figure*}[htbp]
    \centering
    \includegraphics[width=\textwidth]{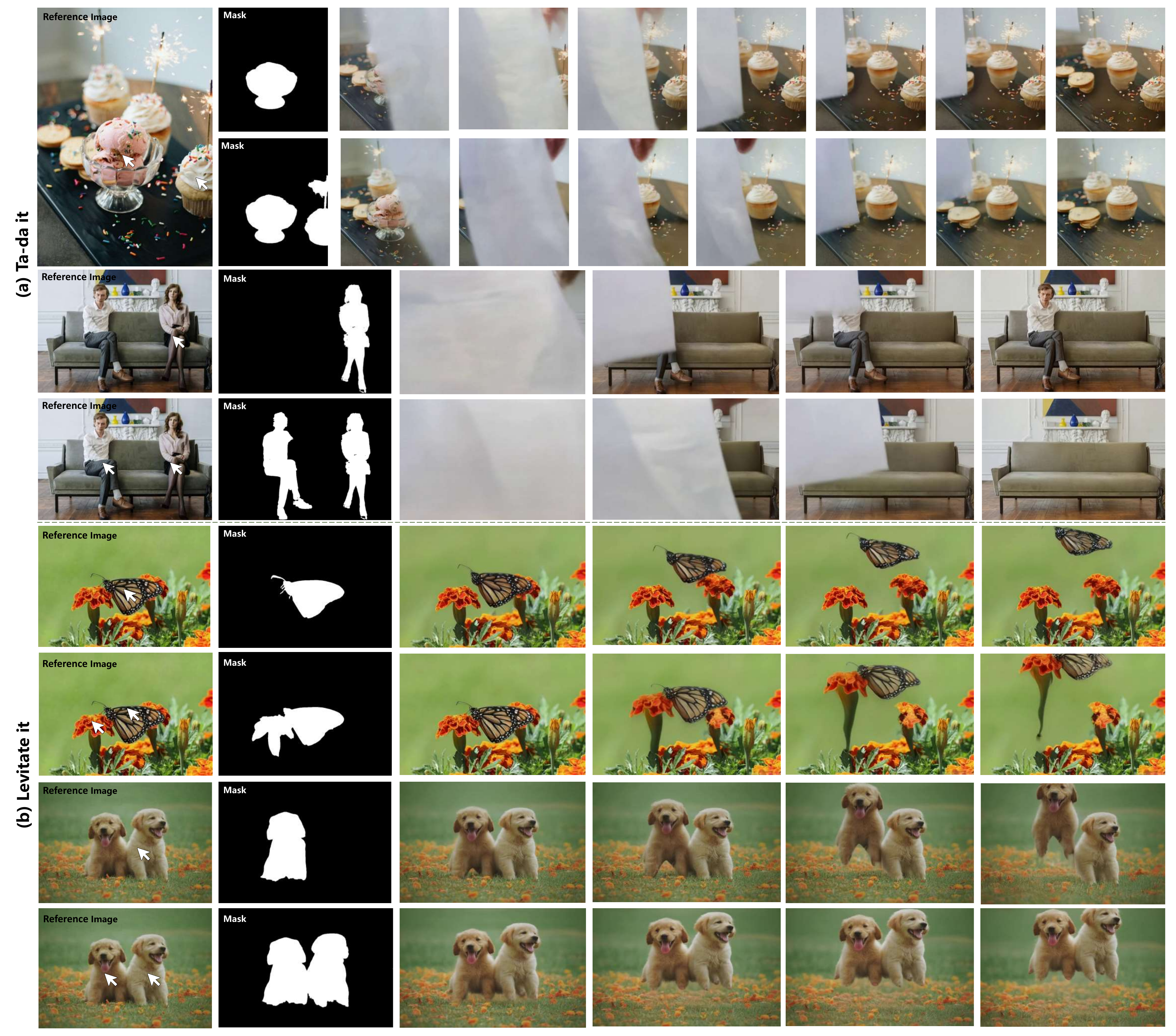}  %
    \caption{More spatial controlled VFX generation results of our method on two different visual effects.}
    \label{fig:control}
\end{figure*}

\begin{figure*}[htbp]
    \centering
    \includegraphics[width=\textwidth]{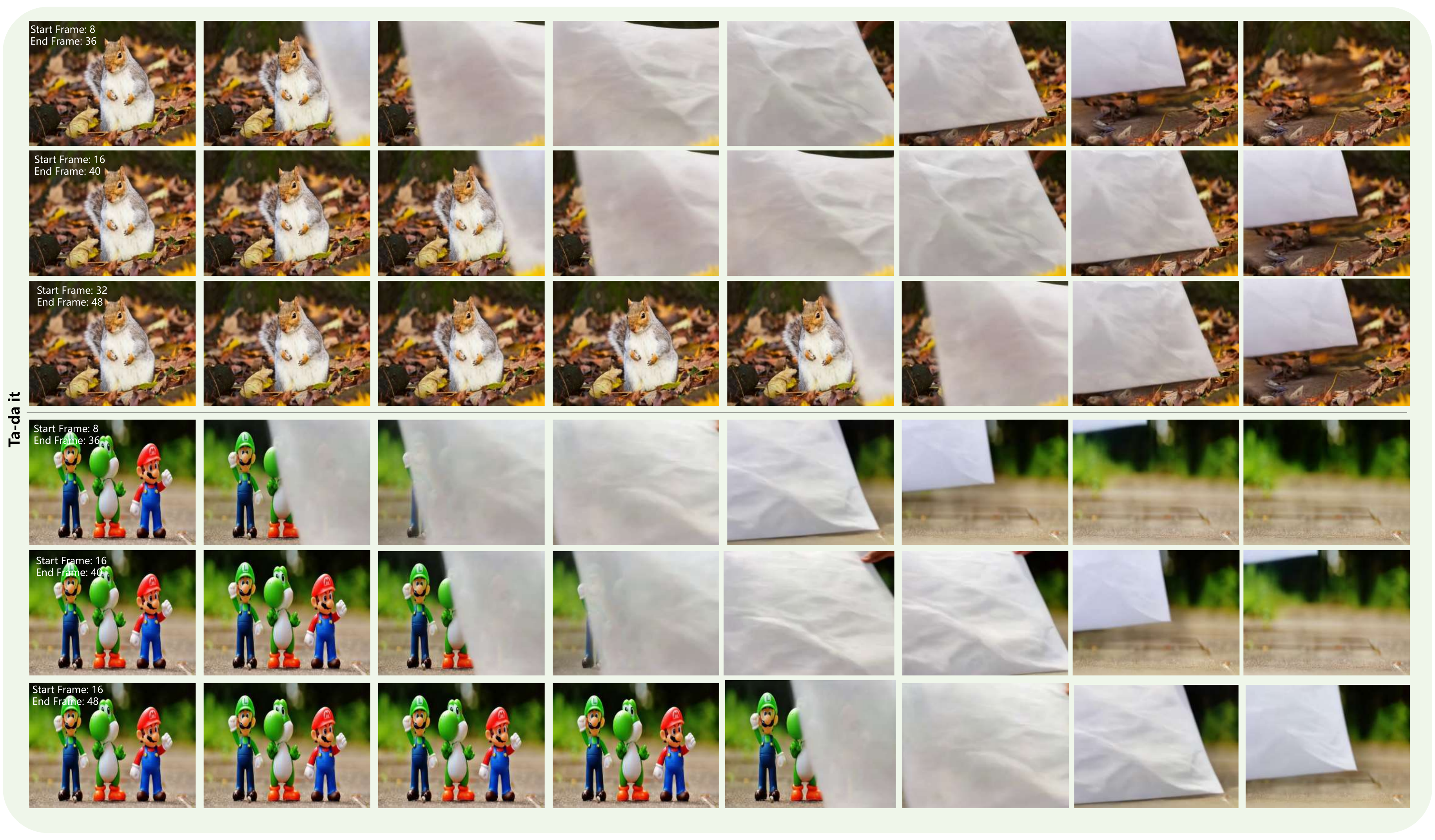}  %
    \caption{More temporal controlled VFX generation results of our method.}
    \label{fig:temp}
\end{figure*}

In this section, we present additional results on spatial and temporal controlled VFX video generation using VFX Creator. Fig.~\ref{fig:control} illustrates qualitative findings for two distinct effects across four cases, showcasing VFX Creator's ability to perform instance-level object manipulation with precision. Additionally, the temporal results depicted in Fig.~\ref{fig:temp} demonstrate VFX Creator's capacity to control the rhythm of VFX video generation over time. This underscores the accuracy and effectiveness of the control mechanisms in our method.
\begin{figure*}[htbp]
    \centering
    \includegraphics[width=\textwidth]{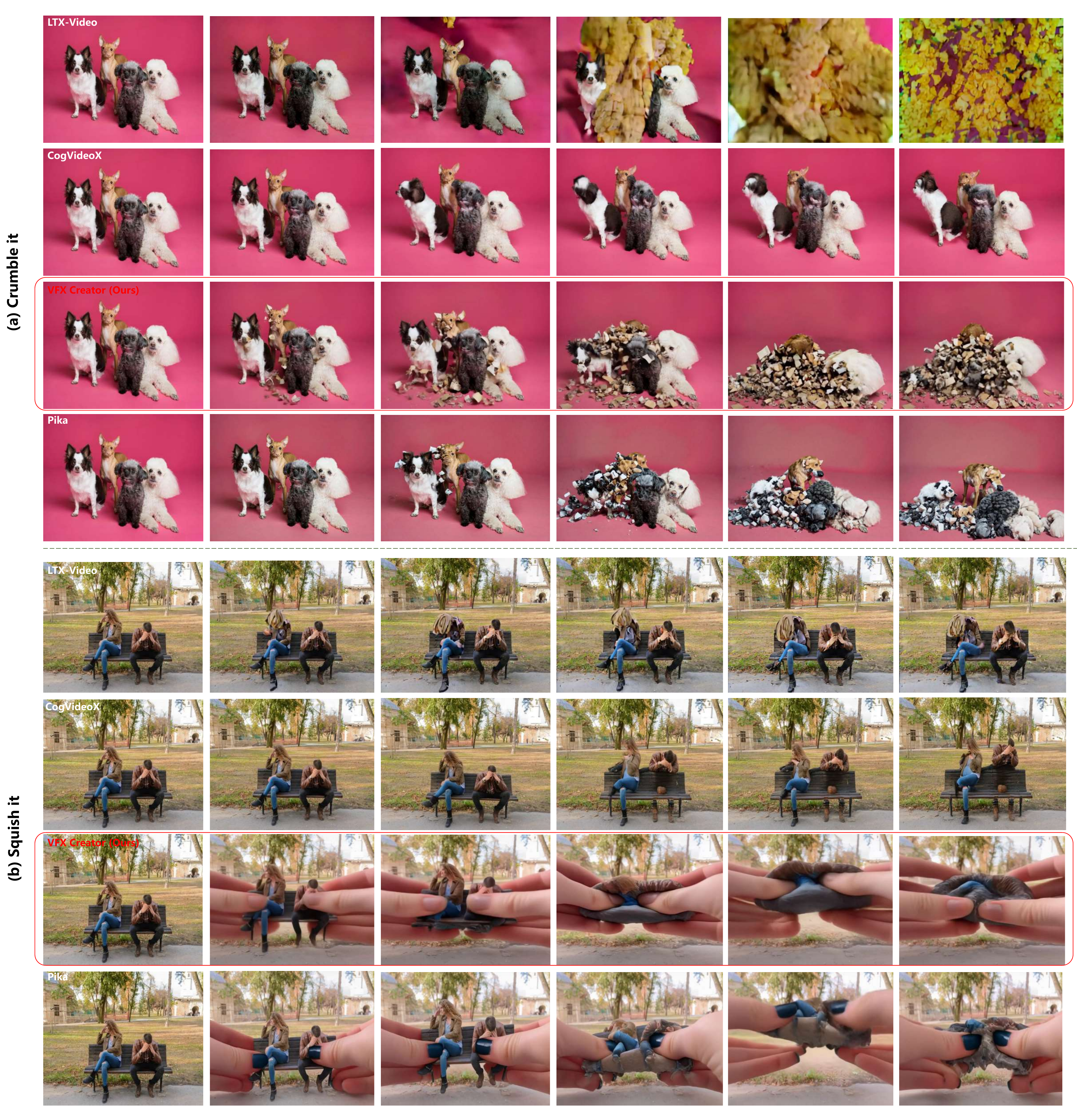}  %
    \caption{More qualitative comparison results of VFX video generation on two different visual effects between our method, DynamiCrafter, LTX-Video, CogVideoX-5B, and Pika.}
    \label{fig:compare_supp}
\end{figure*}
\section{More Qualitative comparison results}
In this section, we present additional qualitative comparison results of VFX Creator with other video generation models, including LTX-Video~\cite{hacohen2024ltx}, and CogVideoX~\cite{yang2024cogvideox}. As illustrated in Fig.~\ref{fig:compare_supp}, LTX-Video and CogVideoX demonstrate a failure to accurately interpret the visual effect prompt, leading to the generation of only minimal actions. In contrast, VFX Creator and Pika (ground truth) exhibit a significantly better comprehension and generation of the corresponding effect videos. Notably, for the "Crumble it" effect, our methodology yields results that are not only more complete but also superior in quality when compared to the ground truth.

\end{document}